\documentclass[11pt]{article}
\usepackage[final]{acl}
\usepackage{times}
\usepackage{latexsym}
\usepackage[T1]{fontenc}
\usepackage[utf8]{inputenc}
\usepackage{microtype}
\usepackage{inconsolata}
\usepackage{graphicx}

\usepackage{tcolorbox}
\usepackage{fontawesome5}
\usepackage{xspace}  
\usepackage{xcolor}  
\usepackage[table,xcdraw]{xcolor}
\usepackage{multicol}
\usepackage{multirow}
\usepackage{booktabs}
\usepackage{pifont}
\usepackage[table]{xcolor}
\definecolor{avggray}{gray}{0.93}
\usepackage{amssymb}
\usepackage{amsmath} 
\usepackage{tabularx}

\newcommand{\cmark}{\ding{51}} 
\newcommand{\xmark}{\ding{55}} 
\definecolor{lightred}{RGB}{255, 230, 230}

\definecolor{mybg}{HTML}{FFFDE9}

\graphicspath{{img/}}   

\title{Rubric-Guided Process Reward for Stepwise Model Routing}

\author{
  \textbf{Shenghao Ye\textnormal{\textsuperscript{1}\footnotemark[1]}},
  \textbf{Yu Guo\textnormal{\textsuperscript{1}\footnotemark[1]}},
  \textbf{Zhengheng Li\textnormal{\textsuperscript{2}\footnotemark[1]}},
  \textbf{Qiang Huang\textnormal{\textsuperscript{5}}},
  \textbf{Shuangwu Chen\textnormal{\textsuperscript{1}\footnotemark[2]}}, 
  \textbf{Wang Luo\textnormal{\textsuperscript{4}}},
\\
  \textbf{Qian Rui\textnormal{\textsuperscript{5}}},
  \textbf{Jie Zhang\textnormal{\textsuperscript{6}}},
  \textbf{Tao Zhang\textnormal{\textsuperscript{1}}},
  \textbf{Dong Jin\textnormal{\textsuperscript{3}\footnotemark[2]}},
  \textbf{Xiaobin Tan\textnormal{\textsuperscript{1}}},
  \textbf{Quan Zheng\textnormal{\textsuperscript{1}}},
  \textbf{Jian Yang\textnormal{\textsuperscript{1}}}
\\
  \textsuperscript{1}University of Science and Technology of China \textsuperscript{2} Southeast University 
\\
  \textsuperscript{3}Institute of Artificial Intelligence, Hefei Comprehensive National Science Center
\\
  \textsuperscript{4} Sun Yat-sen University \textsuperscript{5} Fudan University \textsuperscript{6} TeleAI
\\
  \texttt{\{ssh0321y, yukariguo\}@mail.ustc.edu.cn}
}

\begin{document}
\maketitle
\renewcommand{\thefootnote}{\fnsymbol{footnote}}
\footnotetext[1]{Equal contribution}
\footnotetext[2]{Corresponding authors}
\begin{abstract}
Stepwise model routing improves the efficiency of Large Reasoning Models (LRMs) by assigning each reasoning step to a suitable model. Recent methods formulate routing as a sequential decision process and train the router with reinforcement learning. However, although they model routing as a process, they still supervise the router with outcome rewards. Such rewards only reflect final answer correctness and fail to evaluate intermediate routing decisions, which can weaken performance and generalization. To address this gap, we propose RoRo, a rubric-guided process reward framework for stepwise model routing. RoRo first collects diverse routing trajectories and constructs preference pairs based on outcome, cost, and process quality. It then trains a Rubricor to generate a query-specific evaluation rubric and a Judge to score routing trajectories under this rubric through alternating optimization. The resulting process rewards are combined with outcome rewards to optimize the routing policy via GRPO. Experiments on five reasoning benchmarks under both same-family and cross-family settings show that RoRo consistently outperforms strong baselines and achieves better accuracy and cost trade-offs.
\end{abstract}

\section{Introduction}

Large Reasoning Models (LRMs) \citep{yang2025qwen3,guo2025deepseek} have shown strong performance on complex reasoning tasks, such as mathematical reasoning, logical inference, and multi-hop question answering \citep{zhao2024docmath, newman2024arxivdigestables}. Their success often relies on longer and more explicit reasoning trajectories, which help models break down complex reasoning into simpler intermediate steps. However, this ability comes with a high inference cost. As the reasoning chain becomes longer, both latency and computational overhead increase, making LRMs hard to deploy in latency-sensitive and resource-limited scenarios \citep{snell2025scaling,muennighoff2025s1}.

To address this tension, works \citep{shi2025speccot, pan2025specreason} have explored stepwise model routing, which selects a suitable model for each intermediate reasoning step. Their core idea is that different steps in the same reasoning trajectory can vary substantially in difficulty \citep{ong2025routellm}. In practice, only a few critical steps require the stronger model, while many simpler steps can be handled by the smaller model \citep{wang2025mixllm}. Early methods \citep{zeng2026glimprouter,lee2026confidence} follow this idea by estimating the difficulty of each step independently and using local uncertainty signals to decide when to call the large model.

\begin{figure}[t]
    \centering
    \includegraphics[width=0.89\linewidth]{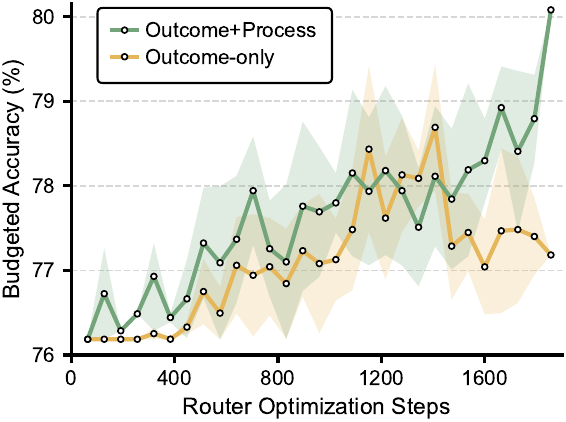}
    \vspace{-8pt}
    \caption{Budgeted accuracy during router optimization on MATH-500 over training. Budgeted accuracy is measured at 60\% of the LRM-only FLOPs.}
    \label{fig:motivation}
    \vspace{-16pt}
\end{figure}

\begin{figure}[t]
    \centering
    \includegraphics[width=0.96\linewidth]{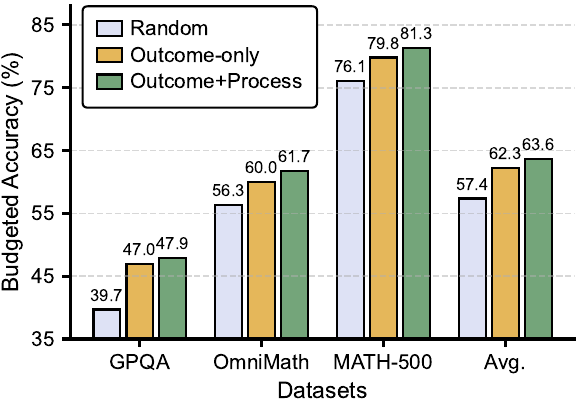}
    \vspace{-8pt}
    \caption{
Budgeted accuracy across MATH-500, OmniMath, and GPQA. Budgeted accuracy is measured at 60\% of the LRM-only FLOPs.
}
    \label{fig:motivation2}
    \vspace{-16pt}
\end{figure}

Despite their promising performance, recent studies \citep{zhang2025routerr1teachingllmsmultiround,kapoor2026trim} further argue that routing should not be viewed as an isolated decision at each step, but as a process over the full reasoning trajectory. From this perspective, the decision to invoke the LRM depends not only on the local difficulty of the current step, but also on the prefix state, such as previous model choices. Following this view, TRIM \citep{kapoor2026trim} formulates routing as a sequential decision process and trains the router with reinforcement learning. It mainly uses final answer correctness as the outcome-level reward. However, these methods still rely on outcome-level supervision, even though they model routing as a process.
Existing studies in reward modeling \citep{wang2026outcomeaccuracyenoughaligning,uesato2023solving} have shown that outcome accuracy alone is insufficient for process modeling. Relying only on final correctness is prone to deceptive alignment, where the model achieves high outcome accuracy while learning misaligned intermediate behaviors. This can weaken both performance and generalization, as shown in Figures~\ref{fig:motivation} and \ref{fig:motivation2}. Therefore, stepwise model routing needs explicit process-level rewards to guide router optimization.

However, routing trajectories are difficult to verify. There are no gold labels that directly indicate whether each routing decision is good or bad, making reliable process rewards hard to obtain. Recent works \citep{sheng2026reinforcing,gunjal2026rubrics} use rubric to evaluate chain-of-thought reasoning and turn these evaluations into process rewards. This idea naturally motivates our study, since both chain-of-thought reasoning and stepwise model routing require process-level evaluation. As shown in Figures~\ref{fig:motivation} and \ref{fig:motivation2}, our preliminary study shows that even simple rubric-based rewards raise the performance ceiling and bring consistent gains across datasets. These results suggest that rubrics can provide useful process supervision for learning better routers.


Building on these findings, we propose \textbf{RoRo}, a \textbf{\underline{r}}ubric-guided pr\textbf{\underline{o}}cess reward framework for stepwise model \textbf{\underline{ro}}uting. RoRo first collects diverse routing trajectories and constructs preference pairs based on outcome, cost, and process quality. It then trains a Rubricor and Judge alternately to learn process-level evaluation for routing. During router optimization, the Rubricor and Judge are frozen. For each query, the Rubricor generates a query-specific rubric, and the Judge scores routing trajectories under this rubric to provide process rewards, which are used to optimize the router with GRPO. In this way, RoRo provides explicit process supervision for learning better routing policies.

\textbf{Our Contributions}.
(1) \underline{\textbf{\textit{New Perspective}}.}
We identify a supervision gap in stepwise model routing, where existing methods model routing as a process but still rely on outcome-only rewards, making them prone to deceptive alignment and limiting performance.
(2) \underline{\textbf{\textit{Novel Framework}}.}
We propose RoRo, a rubric-guided process reward framework that provides process rewards for stepwise routing policy learning.
(3) \underline{\textbf{\textit{SOTA Performance}}.} Experiments on multiple benchmarks show that RoRo consistently outperforms strong baselines and achieves better performance-cost trade-offs.

\begin{figure*}[t]
    \centering
    \includegraphics[width=1\linewidth]{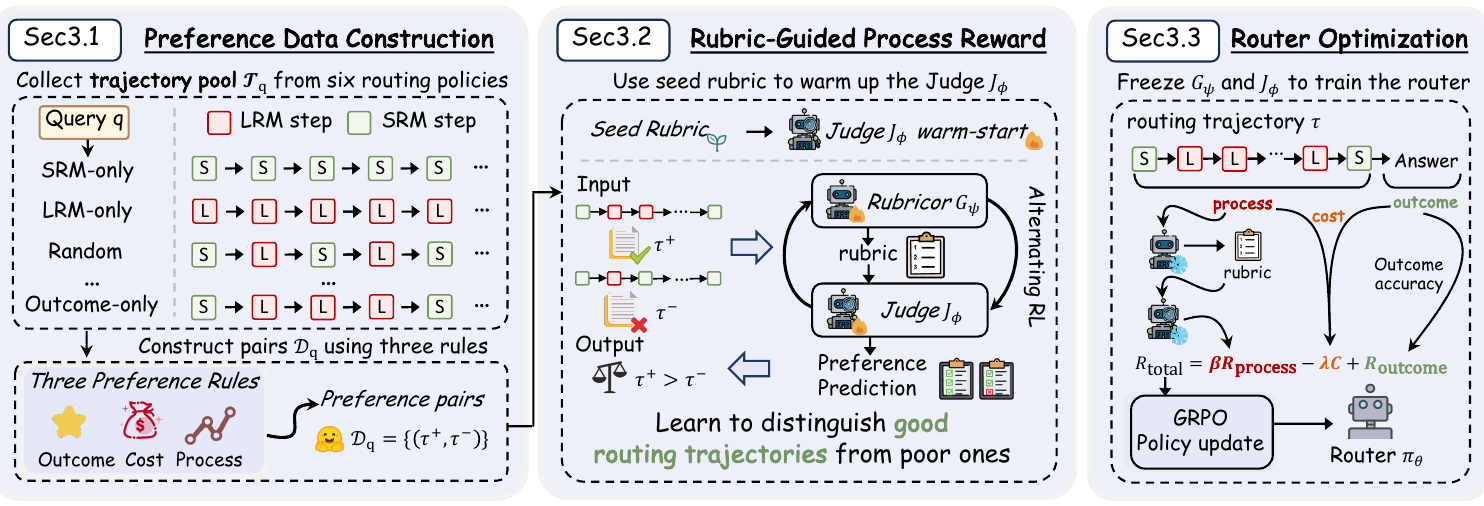}
    \vspace{-22pt}
    \caption{Overview of the RoRo pipeline. Stage 1 constructs route preference data from routing policies. Stage 2 trains the Rubricor and Judge via alternating optimization with validation gating. Stage 3 freezes both components and optimizes the routing policy with GRPO; the Rubricor generates query-specific rubrics.}
    \label{fig:framework}
    \vspace{-14pt}
\end{figure*}


\section{Preliminary}
\label{sec:prelim}

\paragraph{Stepwise Model Routing.}
We formulate stepwise model routing as a sequential decision process. Given a query $q$, a multi-step reasoning task is solved through a sequence of steps $s_1, \ldots, s_n$.  We consider a large reasoning model (LRM) $p_{\theta_M}$, a small reasoning model (SRM) $p_{\theta_m}$. At each step $i$, the SRM first drafts a candidate step $\hat{s}_i$. The routing policy $\pi_\theta$ then chooses an action $a_i \in \{\texttt{continue}, \texttt{regenerate}\}$. If $a_i = \texttt{continue}$, the draft is accepted as $s_i = \hat{s}_i$. If $a_i = \texttt{regenerate}$, the LRM regenerates the step with higher reliability and higher cost. We define a \textit{routing trajectory} as $\tau = (s_{1:n}, a_{1:n})$, which contains the reasoning steps $s_{1:n}$ and routing actions $a_{1:n}$. In this work, we split reasoning traces into steps using the newline delimiter ``\texttt{\textbackslash n\textbackslash n}'' \citep{lee2026confidence}. The routing policy $\pi_\theta$ takes the candidate step $\hat{s}_i$ and a prefix summary as input. The prefix summary contains the query $q$, previous per-step routing actions, per-step costs, and per-step uncertainty signals of previous steps. We use average token entropy as the default uncertainty signal \citep{zeng2026glimprouter}.


\paragraph{Rubric-Based Process Reward.}
Rubric-based reward modeling \citep{liu2025openrubrics,gunjal2026rubrics} evaluates non-verifiable processes with explicit criteria. A rubric $r = \{(c_i, w_i)\}_{i=1}^{|r|}$ contains natural-language criteria. Each criterion $c_i$ describes one aspect, and $w_i \in [0,1]$ denotes its weight. In our setting, a rubric evaluates the quality of routing trajectories. Given a query $q$ and a trajectory set $\mathcal{T}_q = \{\tau^{(j)}\}_{j=1}^N$, a rubric generator $G_\psi$ produces a query-specific rubric
$
r_q = G_\psi(q,\, \mathcal{T}_q).
$
A judge $J_\phi$ scores each trajectory $\tau \in \mathcal{T}_q$ under this rubric and produces a process reward
$
R_{\mathrm{proc}} = J_\phi(q,\, \tau,\, r_q).
$
This reward captures routing quality beyond final-answer correctness.

\section{RoRo}
\label{sec:method}

We present RoRo, a \textbf{\underline{r}}ubric-guided pr\textbf{\underline{o}}cess reward framework for stepwise model \textbf{\underline{ro}}uting. The overall training pipeline consists of three stages, as shown in Figure~\ref{fig:framework}. First, we collect routing trajectories from diverse policies and construct preference pairs based on outcome, cost, and process quality (Sec.~\ref{sec:trajectory_collection}). Second, we train a Rubricor $G_\psi$ and a Judge $J_\phi$ through alternating optimization (Sec.~\ref{sec:rubric_reward_modeling}). For each query, $G_\psi$ generates a query-specific rubric, and $J_\phi$ scores routing trajectories under this rubric. Third, we freeze both components and optimize the routing policy with GRPO, using the process rewards produced by the trained $G_\psi$ and $J_\phi$ during policy learning (Sec.~\ref{sec:router_optimization}).

\subsection{Route Preference Data Construction}
\label{sec:trajectory_collection}

\paragraph{Trajectory Collection.}
For each training query $q$, we collect routing trajectories from six policies to ensure diversity. (1) \textit{SRM-only}, which assigns every step to the small model; (2) \textit{LRM-only}, which assigns every step to the large model; (3) \textit{Random routing}, which assigns each step to either model with a random probability; (4) \textit{Entropy-based routing} \citep{zeng2026glimprouter}, which routes to the LRM when step entropy exceeds a threshold; (5) \textit{Confidence-based routing} \citep{lee2026confidence}, which routes to the LRM when model confidence is low; and (6) \textit{Outcome-only router}, a policy trained with only outcome reward and cost penalty. The first five policies are training-free. Together, these policies span diverse routing behaviors and cost-performance trade-offs. We denote the collected trajectory pool for query $q$ as $\mathcal{T}_q = \{\tau^{(j)}\}_{j=1}^{N_q}$.

\paragraph{Preference Pair Construction.}
From the trajectory pool $\mathcal{T}_q$, we construct preference pairs $(\tau^+, \tau^-)$, where $\tau^+ \succ \tau^-$ indicates that $\tau^+$ is preferred to $\tau^-$. We define preferences using three rules.
(1) \textit{Outcome preference} favors the trajectory with a correct final answer. (2) \textit{Cost preference} favors the trajectory with lower LRM usage when both answers are correct. (3) \textit{Process preference} favors the trajectory with better routing quality when outcome and cost are similar. The first two criteria are computed from final answers and token counts. For process preference, we use the seed rubric described in Sec.~\ref{sec:rubric_reward_modeling} to score each trajectory and select the one with the higher rubric score. We denote preference set for query $q$ as $\mathcal{D}_q$.

\subsection{Rubric-Guided Process Reward Modeling}
\label{sec:rubric_reward_modeling}

Since the routing process has no gold labels, we cannot directly supervise each routing decision \citep{guha2024smoothie}. To address this, we train a Rubricor $G_\psi$ and a Judge $J_\phi$ to learn what makes a routing trajectory good. For each query, $G_\psi$ generates a rubric from the trajectory pool, and $J_\phi$ scores each trajectory under this rubric.

\paragraph{Seed Rubric.}
We manually design a seed rubric $r_{\mathrm{seed}}$ that captures basic routing quality criteria, as detailed in Appendix~\ref{sec:appendix_rubrics}. The seed rubric contains multiple criteria with uniform weights. Each criterion maps trajectory features to a normalized score in $[0,1]$. It serves two roles. First, it conditions the Judge during initialization. Second, it provides initial process preferences for constructing $\mathcal{D}_q$.

\paragraph{Judge Warm-Start.}
We first warm up the Judge using the seed rubric. For each query $q$ and preference pair $(\tau^+, \tau^-) \in \mathcal{D}_q$, we train $J_\phi$ with the Bradley-Terry objective
\begin{equation}
\vspace{-2pt}
\begin{aligned}
\mathcal{L}_J = -\log \sigma\bigl(
&J_\phi(q, \tau^+, r_{\mathrm{seed}}) \\
&- J_\phi(q, \tau^-, r_{\mathrm{seed}})\bigr),
\end{aligned}
\end{equation}
where $J_\phi(q, \tau, r)$ denotes the scalar score assigned by the Judge to trajectory $\tau$ under rubric $r$ for query $q$, and $\sigma$ is the sigmoid function. Here, $\tau^+$ and $\tau^-$ denote the preferred and dispreferred trajectories. This stage gives the Judge a basic ability to rank routing trajectories under a rubric. The Judge is further improved during alternating optimization.


\paragraph{Alternating Optimization.}
After warm-start, we refine the Rubricor and the Judge over multiple rounds, indexed by $t$.

\textbf{Phase 1: Update Rubricor.} We freeze $J_\phi^{(t)}$ and optimize $G_\psi$. For each query $q$, the Rubricor takes the trajectory pool $\mathcal{T}_q$ as input and samples $M$ candidate rubrics $\{r_{q,m}\}_{m=1}^M$. 
Each candidate rubric $r_{q,m} = \{(c_i, w_i)\}$ contains criteria with weights.

Before updating the Rubricor, we validate the criteria in each candidate rubric on a held out set of routing rollouts. For each criterion $c_i$, we compute its partial correlation with route preference after controlling for outcome and cost. A criterion is retained only if (1) the partial correlation is statistically significant, (2) its score variance across trajectories exceeds a minimum threshold, and (3) the mutual information between its score and the outcome label stays below a leakage threshold. The retained criteria and their weights form a validated candidate rubric $\tilde{r}_{q,m}$. This filtering procedure follows RLCER~\citep{sheng2026reinforcing}, adapted for routing where outcome leakage should be avoided. We provide full details of the validation gate in Appendix~\ref{sec:appendix_validation}.
Each validated candidate rubric receives a reward based on its average preference margin under the frozen Judge:
\[
\rho_{q,m}
=
\log \sigma
\left(
\bar{\Delta}J^{(t)}_{q,m}
\right),
\]
where $\bar{\Delta}J^{(t)}_{q,m}$ is the mean of
$
J^{(t)}_\phi(q,\tau^+,\tilde r_{q,m})
-
J^{(t)}_\phi(q,\tau^-,\tilde r_{q,m})
$
over all preference pairs $(\tau^+,\tau^-)\in \mathcal{D}_q$.
A larger reward indicates that the generated rubric helps the Judge better separate preferred trajectories from dispreferred ones.
We then treat $\rho_{q,m}$ as the scalar reward for the sampled rubric $ r_{q,m}$ and update the Rubricor with policy-gradient optimization:
\[
\nabla_\psi \mathcal{J}_G
=
\mathbb{E}_{q,m}
\left[
\rho_{q,m}
\nabla_\psi
\log G_\psi(r_{q,m}\mid q,\mathcal{T}_q)
\right].
\]
Thus, the Rubricor learns to generate rubrics under which the Judge assigns higher scores to preferred trajectories than to dispreferred ones.


\textbf{Phase 2: Update Judge.} We freeze the updated Rubricor and optimize $J_\phi$. For each query $q$, the Rubricor generates a shared rubric from the trajectory pool. The same validation step is applied to obtain a validated rubric $\tilde{r}_q^{(t)}$. The Judge is then trained on all pairs within that query group
\begin{equation}
\begin{aligned}
\mathcal{L}_J^{(t)} = -\sum_{q}\sum_{(\tau^+, \tau^-) \in \mathcal{D}_q} \log \sigma\bigl(
&J_\phi(q, \tau^+, \tilde{r}_q^{(t)}) \\
&- J_\phi(q, \tau^-, \tilde{r}_q^{(t)})\bigr).
\end{aligned}
\end{equation}

Through this alternating process, the Rubricor learns to generate valid criteria for routing evaluation, while the Judge learns to score routing trajectories under these criteria. Together, they produce reliable process rewards for router training.

\paragraph{Process Reward Computation.}
During router training, we freeze the Rubricor and the Judge. Given a query $q$ and its sampled GRPO rollout group $\mathcal{T}_q = \{\tau^{(k)}\}_{k=1}^{K}$, the Rubricor first generates a candidate rubric
\begin{equation}
r_q = G_\psi(q,\, \mathcal{T}_q).
\end{equation}
We then apply the validation gate to $r_q$. If $r_q$ fails the validation gate, the rollout group is discarded. Otherwise, the validated rubric $\tilde{r}_q$ is used to score each trajectory
\begin{equation}
R_{\mathrm{process}}^{(k)} = J_\phi(q,\, \tau^{(k)},\, \tilde{r}_q).
\end{equation}
Since all trajectories in the same rollout group share $\tilde{r}_q$, their process rewards are directly comparable.

\subsection{Process-Aware Router Optimization}
\label{sec:router_optimization}

With $G_\psi$ and $J_\phi$ frozen, we optimize the routing policy $\pi_\theta$ using GRPO \citep{shao2024deepseekmathpushinglimitsmathematical}.

\paragraph{Training Objective.}
For each query $q$, the router samples a group of $K$ routing trajectories $\{\tau^{(k)}\}_{k=1}^{K}$. Following the process reward computation above, the Rubricor generates a validated rubric $\tilde{r}_q$ for this group, and the Judge assigns a process reward to each trajectory. Each trajectory receives the combined reward
\begin{equation}
R_{\mathrm{total}}^{(k)} = R_{\mathrm{outcome}}^{(k)} - \lambda C^{(k)} + \beta R_{\mathrm{process}}^{(k)},
\end{equation}
where $R_{\mathrm{outcome}}^{(k)} \in \{0,1\}$ indicates answer correctness, $C^{(k)}$ counts LRM generation tokens, and $R_{\mathrm{process}}^{(k)}$ is the process reward from the Judge. $\lambda$ and $\beta$ control the weights of cost and quality.

\paragraph{Advantage Estimation.}
GRPO computes advantages by normalizing rewards within each rollout group
\begin{equation}
A^{(k)} = \frac{R_{\mathrm{total}}^{(k)} - \mu_K}{\sigma_K},
\end{equation}
where $\mu_K$ and $\sigma_K$ are the group reward mean and standard deviation. This removes the separate value model. Each decision in trajectory $\tau^{(k)}$ shares the same trajectory-level advantage $A^{(k)}$. We then update the router with the GRPO objective using $A^{(k)}$ as the trajectory-level advantage.

\subsection{Overall Inference Pipeline}
Given a query $q$, RoRo generates an answer through iterative routing. At each step, the SRM drafts a candidate step, and the router decides whether to accept it or invoke the LRM.

(1) \textit{Initial draft}. The SRM $p_{\theta_m}$ generates the first candidate step $\hat{s}_1$.

(2) \textit{Routing decision}. The router $\pi_\theta$ takes the candidate step $\hat{s}_i$ and a compact prefix summary as input. The summary records the query $q$, previous actions, per-step costs, and the average token entropy of each selected step, where the entropy comes from the model that produced that step. Based on this input, the router selects an action $a_i \in \{\texttt{continue}, \texttt{regenerate}\}$.

(3) \textit{Step update}. If $a_i = \texttt{continue}$, the candidate step is accepted as $s_i = \hat{s}_i$. If $a_i = \texttt{regenerate}$, the LRM $p_{\theta_M}$ regenerates this step, and the regenerated step is used as $s_i$.

(4) \textit{Iterative generation}. After obtaining $s_i$, RoRo appends it to the current reasoning trace. The SRM then drafts the next candidate step $\hat{s}_{i+1}$, and the same routing process is repeated.

(5) \textit{Answer completion}. The process stops when the model produces a complete final answer or reaches the predefined maximum step limit.

\section{Experiments}
\label{sec:exp}
We present a comprehensive evaluation of RoRo. We first describe the experimental setup (Sec.~\ref{sec:exp_setup}). We then organize our experiments around four key research questions.
\textbf{Q1}: Does RoRo consistently outperform existing stepwise routing methods? (Sec.~\ref{sec:exp_main})
\textbf{Q2}: How does each component contribute to RoRo's performance? (Sec.~\ref{sec:exp_ablation})
\textbf{Q3}: Does RoRo provide better cost-effectiveness under different inference budgets? (Sec.~\ref{sec:exp_robust})
\textbf{Q4}: Is RoRo sensitive to problem difficulty and reliable across uncertainty signal choices? (Sec.~\ref{sec:exp_analysis})

\begin{table*}[t]
  \centering
  \resizebox{\textwidth}{!}{
  \begin{tabular}{l *{15}{c}|>{\columncolor{avggray}}c>{\columncolor{avggray}}c>{\columncolor{avggray}}c}
  \toprule
  \multirow{3}{*}{\textbf{Method}} 
  & \multicolumn{9}{c}{\textbf{In-domain}} 
  & \multicolumn{6}{c}{\textbf{Out-of-domain}} 
  & \multicolumn{3}{c}{\cellcolor{avggray}\textbf{Avg.}} \\
  \cmidrule(lr){2-10} \cmidrule(lr){11-16} \cmidrule(lr){17-19}
  & \multicolumn{3}{c}{\textbf{MATH-500}} 
  & \multicolumn{3}{c}{\textbf{AIME 2025}} 
  & \multicolumn{3}{c}{\textbf{OmniMath}} 
  & \multicolumn{3}{c}{\textbf{GSM8K}} 
  & \multicolumn{3}{c}{\textbf{GPQA}} 
  & \multicolumn{3}{c}{\cellcolor{avggray}\textbf{Average}} \\
  \cmidrule(lr){2-4} 
  \cmidrule(lr){5-7} 
  \cmidrule(lr){8-10} 
  \cmidrule(lr){11-13} 
  \cmidrule(lr){14-16} 
  \cmidrule(lr){17-19}
  & @20 & @40 & @60 
  & @20 & @40 & @60 
  & @20 & @40 & @60 
  & @20 & @40 & @60 
  & @20 & @40 & @60 
  & @20 & @40 & @60 \\
  \midrule
  \multicolumn{19}{c}{\textbf{Qwen3-1.7B / Qwen3-14B (Same-family)}}\\
  \midrule\midrule
  SRM-only  
  &  & 68.6 &  
  &  & 32.4 & 
  &  & 51.5 &  
  &  & 87.3 &  
  &  & 32.8 &  
  & -- & -- & -- \\
  
  LRM-only  
  &  & 86.2 &  
  &  & 66.6 &  
  &  & 63.4 &  
  &  & 97.0 &  
  &  & 52.3 &  
  & -- & -- & -- \\
  
  \midrule
  Random        
  & 70.4 & 74.2 & 76.1 
  & 35.4 & 43.1 & 50.8 
  & 52.7 & 55.0 & 56.3 
  & 88.3 & 90.3 & 92.3 
  & 33.4 & 36.5 & 39.7 
  & 56.0 & 59.8 & 63.4 \\
  
  RSD           
  & 72.8 & 75.8 & 79.2 
  & 36.9 & 45.8 & 53.7 
  & 54.7 & 57.6 & 59.8 
  & 89.5 & 93.5 & 95.3 
  & 35.7 & 41.6 & 46.2 
  & 57.9 & 62.9 & 66.8 \\
  
  SpecCoT       
  & 73.4 & 77.8 & 78.8 
  & 37.3 & 44.9 & 59.4 
  & 52.2 & 56.4 & 58.3 
  & 89.8 & 91.5 & 93.7 
  & 34.2 & 38.1 & 42.5 
  & 57.4 & 61.7 & 66.5 \\
  
  SpecReason    
  & 73.9 & 76.9 & 79.4 
  & 37.6 & 45.5 & 56.2 
  & \textbf{57.0} & \underline{58.2} & 60.1 
  & 90.3 & 92.4 & 94.2 
  & 34.6 & 39.0 & 43.3 
  & 58.7 & 62.4 & 66.6 \\
  
  STEER         
  & 74.3 & 77.4 & 79.9 
  & 37.1 & 45.9 & 57.5 
  & 54.1 & 57.3 & \underline{62.5} 
  & 91.2 & 93.6 & 95.0 
  & 35.1 & 40.5 & 44.8 
  & 58.4 & 62.9 & 67.3 \\
  
  GlimpRouter   
  & \textbf{76.0} & \underline{78.0} & \underline{80.1} 
  & \underline{38.2} & \textbf{47.6} & 58.8 
  & 55.2 & 57.5 & 60.3
  & 92.3 & 94.2 & 95.4 
  & 35.8 & 42.0 & 46.5 
  & \underline{59.5} & \underline{63.9} & 68.2 \\
  
  TRIM          
  & 75.1 & 77.5 & 79.8 
  & 34.8 & 45.3 & \underline{60.3} 
  & 55.9 & 57.2 & 60.0 
  & \underline{94.1} & \underline{95.5} & \underline{95.8} 
  & \underline{36.4} & \underline{43.6} & \underline{47.0} 
  & 59.3 & 63.8 & \underline{68.6} \\
  
  \rowcolor[HTML]{D8ECE4}
  \textbf{RoRo (Ours)} 
  & \underline{75.5} & \textbf{79.8} & \textbf{81.9} 
  & \textbf{40.6} & \underline{47.0} & \textbf{60.5} 
  & \underline{56.5} & \textbf{59.9} & \textbf{62.8} 
  & \textbf{95.2} & \textbf{96.1} & \textbf{96.5} 
  & \textbf{37.2} & \textbf{44.6} & \textbf{48.6} 
  & \textbf{61.0} & \textbf{65.5} & \textbf{70.0} \\
  \midrule
  \multicolumn{19}{c}{\textbf{Qwen3-1.7B / DeepSeek-R1-Distill-Qwen-14B (Cross-family)}}\\
  \midrule\midrule
  SRM-only  
  &  & 68.6 &  
  &  & 32.4 & 
  &  & 51.5 &  
  &  & 87.3 &  
  &  & 32.8 &  
  & -- & -- & -- \\
  
  LRM-only  
  &  & 82.6 &  
  &  & 65.1 &  
  &  & 62.7 &  
  &  & 94.8 &  
  &  & 48.6 &  
  & -- & -- & -- \\
  
  \midrule
  Random        
  & 69.8 & 73.5 & 75.4 
  & 34.6 & 42.0 & 49.5 
  & 52.0 & 54.3 & 56.6 
  & 87.9 & 89.8 & 91.6 
  & 30.2 & 33.8 & 37.1 
  & 54.9 & 58.7 & 62.2 \\
  
  RSD           
  & 71.5 & 74.3 & 77.8 
  & 36.0 & 44.3 & 52.2 
  & 53.4 & 56.1 & 58.5 
  & 88.7 & 91.8 & 93.9 
  & 32.4 & 37.6 & 42.3 
  & 56.4 & 60.8 & 64.9 \\
  
  SpecCoT       
  & 72.1 & 75.8 & 78.0 
  & 36.4 & 43.5 & 56.8 
  & 51.6 & 55.4 & 57.2 
  & 89.2 & 91.2 & 93.5 
  & 31.5 & 36.0 & 41.2 
  & 56.2 & 60.4 & 65.3 \\
  
  SpecReason    
  & 72.6 & 75.2 & 78.4 
  & 37.0 & 44.6 & 53.8 
  & 52.8 & 55.7 & 58.0 
  & 89.5 & 91.9 & 93.5 
  & \textbf{35.3} & 37.0 & 41.6 
  & 57.4 & 60.9 & 65.1 \\
  
  STEER         
  & 73.1 & 75.8 & 78.8 
  & 36.5 & 44.2 & 55.9 
  & 53.2 & 56.2 & 58.6 
  & 90.1 & 92.3 & 93.8 
  & 32.6 & 38.0 & 42.5 
  & 57.1 & 61.3 & 65.9 \\
  
  GlimpRouter   
  & 73.4 & \textbf{78.8} & \underline{79.2} 
  & \textbf{39.6} & \underline{45.0} & 57.0 
  & 54.0 & \underline{56.6} & \underline{59.0} 
  & 90.9 & 92.7 & 93.8 
  & 33.0 & 39.2 & 43.2 
  & \underline{58.2} & \underline{62.5} & 66.4 \\
  
  TRIM          
  & \underline{73.8} & 76.2 & 78.5 
  & 34.0 & 44.0 & \underline{58.2} 
  & \underline{54.6} & 56.3 & 58.8 
  & \underline{91.6} & \underline{93.1} & \underline{94.0} 
  & 33.4 & \underline{40.3} & \underline{44.0} 
  & 57.5 & 62.0 & \underline{66.7} \\
  
  \rowcolor[HTML]{D8ECE4}
  \textbf{RoRo (Ours)} 
  & \textbf{74.2} & \underline{78.2} & \textbf{80.6} 
  & \underline{38.9} & \textbf{45.7} & \textbf{59.0} 
  & \textbf{55.3} & \textbf{58.5} & \textbf{61.1} 
  & \textbf{92.4} & \textbf{93.6} & \textbf{94.2} 
  & \underline{34.7} & \textbf{41.8} & \textbf{46.1} 
  & \textbf{59.1} & \textbf{63.6} & \textbf{68.2} \\
  
  \bottomrule
  \end{tabular}%
  }
  \caption{
  Main results on five benchmarks under same-family and cross-family settings. We report Budgeted Accuracy (BA) at 20\%, 40\%, and 60\% of LRM-only FLOPs, denoted as @20, @40, and @60. Higher is better. MATH-500, AIME 2025, and OmniMath are in-domain math reasoning benchmarks. GSM8K and GPQA are out-of-domain benchmarks for testing generalization. The best result is \textbf{bold} and the second best is \underline{underlined}.
  }
  \label{tab:main}
  \vspace{-10pt}
  \end{table*}

\subsection{Experimental Setup}
\label{sec:exp_setup}

\begin{table}[t]
\centering
\resizebox{0.48\textwidth}{!}{
\begin{tabular}{l|cc|cc}
\toprule
\textbf{Variant} & \textbf{MATH-500} & $\triangledown$ & \textbf{OmniMath} & $\triangledown$ \\
\midrule
\textbf{RoRo (Qwen3)} & \textbf{79.3} & -- & \textbf{59.9} & -- \\
\quad w/o process reward & 77.1 & \textcolor[HTML]{CC0000}{(-2.2)} & 56.5 & \textcolor[HTML]{CC0000}{(-3.4)} \\
\quad w/o Learned Rubricor & 77.8 & \textcolor[HTML]{CC0000}{(-1.5)} & 58.4 & \textcolor[HTML]{CC0000}{(-1.5)} \\
\quad w/o Learned Judge & 78.2 & \textcolor[HTML]{CC0000}{(-1.1)} & 58.7 & \textcolor[HTML]{CC0000}{(-1.2)} \\
\quad w/o validation & 78.5 & \textcolor[HTML]{CC0000}{(-0.8)} & 58.1 & \textcolor[HTML]{CC0000}{(-1.8)} \\
\midrule
\textbf{RoRo (DeepSeek-R1)} & \textbf{78.0} & -- & \textbf{58.6} & -- \\
\quad w/o process reward & 75.5 & \textcolor[HTML]{CC0000}{(-2.5)} & 55.2 & \textcolor[HTML]{CC0000}{(-3.4)} \\
\quad w/o Learned Rubricor & 76.4 & \textcolor[HTML]{CC0000}{(-1.6)} & 57.0 & \textcolor[HTML]{CC0000}{(-1.6)} \\
\quad w/o Learned Judge & 76.8 & \textcolor[HTML]{CC0000}{(-1.2)} & 57.3 & \textcolor[HTML]{CC0000}{(-1.3)} \\
\quad w/o validation & 77.2 & \textcolor[HTML]{CC0000}{(-0.8)} & 56.8 & \textcolor[HTML]{CC0000}{(-1.8)} \\
\bottomrule
\end{tabular}
}
\vspace{-6pt}
\caption{Ablation results of RoRo. We report average BA across @20, @40, and @60.}
\label{tab:ablation}
\vspace{-14pt}
\end{table}

\paragraph{Models and Configurations.}
We use Qwen3-1.7B \citep{yang2025qwen3} as the SRM and evaluate two LRMs: Qwen3-14B \citep{yang2025qwen3} and DeepSeek-R1-Distill-Qwen-14B \citep{guo2025deepseek}. This setup allows us to study both same-family and cross-family collaboration. All models run in thinking mode for multi-step reasoning. The routing policy is a 2-layer MLP with 128 hidden units, following TRIM \citep{kapoor2026trim}. For the Rubricor and Judge, we both use Qwen3-8B \citep{yang2025qwen3} as the backbone model. The router, Rubricor, and Judge are all trained on the MATH training set \citep{hendrycks2021measuring}.

\paragraph{Benchmarks.}
We evaluate on five benchmarks spanning domains and difficulty levels. MATH-500 \citep{lightman2023let}, AIME 2025 \citep{dekoninck2026beyond}, and OmniMath \citep{gao2025omnimath} serve as in-domain mathematical reasoning benchmarks. To test cross-domain generalization, we include GSM8K \citep{cobbe2021training} and GPQA \citep{rein2024gpqa}. GSM8K contains grade-school math problems that are simpler than the training data, while GPQA covers graduate-level science questions that require reasoning beyond mathematical domain.

\paragraph{Baselines.}
We compare RoRo against standalone models (SRM-only and LRM-only), Random routing, and six representative stepwise routing methods, including RSD \citep{liao2025rewardguided}, SpecCoT \citep{shi2025speccot}, SpecReason \citep{pan2025specreason}, STEER \citep{lee2026confidence}, GlimpRouter \citep{zeng2026glimprouter}, and TRIM \citep{kapoor2026trim}.

\paragraph{Evaluation Metric.}
Following prior work \citep{liao2025rewardguided,sardana2024beyond}, we estimate inference cost using the standard Transformer FLOPs approximation of $2N$ per generated token for a model with $N$ parameters. We report Budgeted Accuracy (BA) at 20\%, 40\%, and 60\% of LRM-only FLOPs, denoted as BA@20, BA@40, and BA@60. For each method, we sweep the routing threshold with a step size of 0.05 and report the highest accuracy within each budget.

\subsection{Main Results}
\label{sec:exp_main}

Table~\ref{tab:main} presents the main results on five benchmarks under both same-family and cross-family settings. Overall, RoRo achieves the highest average BA across all budget levels in both settings.

\paragraph{Same-Family Analysis.}
Under the same-family setting (Qwen3-1.7B / Qwen3-14B), RoRo ranks first in 12 of 15 settings and second in the remaining 3, where GlimpRouter or SpecReason occasionally leads on specific benchmarks at tight budgets. Notably, RoRo achieves the best average BA at all three budget levels. Compared with TRIM, the strongest RL-based baseline using outcome-only reward, RoRo improves average BA@20 by 1.9 points, confirming that rubric-guided process reward provides meaningful additional supervision beyond outcome and cost. On the out-of-domain benchmarks GSM8K and GPQA, RoRo outperforms all baselines despite being trained exclusively on MATH, suggesting that the rubric-guided routing criteria capture general routing principles that transfer across domains and difficulty levels.

\paragraph{Cross-Family Analysis.}
Under the cross-family setting (Qwen3-1.7B / DeepSeek-R1-Distill-Qwen-14B), RoRo again achieves the highest average BA and ranks first in 12 of 15 settings. GlimpRouter leads on AIME@20 and MATH@40, and SpecReason on GPQA@20, but RoRo maintains a clear average advantage. This result verifies that the benefit of RoRo is not limited to a specific model family but generalizes to cross-family collaboration. Another notable advantage of RoRo is that it introduces no extra token generation during routing. Unlike RSD and SpecReason that require additional model calls during inference, RoRo uses only the lightweight MLP router after training.

\subsection{Ablation Study}
\label{sec:exp_ablation}

Table~\ref{tab:ablation} reports ablation results on MATH-500 and OmniMath. The full RoRo consistently achieves the best performance across all budget levels and both benchmarks. Removing the process reward entirely (\emph{w/o process reward}) causes the largest drop, reducing the system to outcome-only training similar to TRIM. This confirms that process reward is the most critical contribution of RoRo. Replacing the learned Rubricor with the fixed seed rubric (\emph{w/o Learned Rubricor}) also degrades performance, showing that the Rubricor discovers rubric dimensions beyond the initial human design. Similarly, removing the learned Judge (\emph{w/o Learned Judge}) and using raw rubric scores hurts accuracy, indicating that the Judge produces more calibrated process rewards than direct scoring. Finally, skipping rubric validation (\emph{w/o validation}) also leads to a clear drop. This suggests that some generated criteria are noisy or redundant, and statistical validation is necessary to retain only genuinely predictive ones. Overall, all components contribute to RoRo, with the process reward itself providing the largest single improvement.

\subsection{Cost-Effectiveness}
\label{sec:exp_robust}

\begin{figure}[t]
    \centering
    \includegraphics[width=0.96\linewidth]{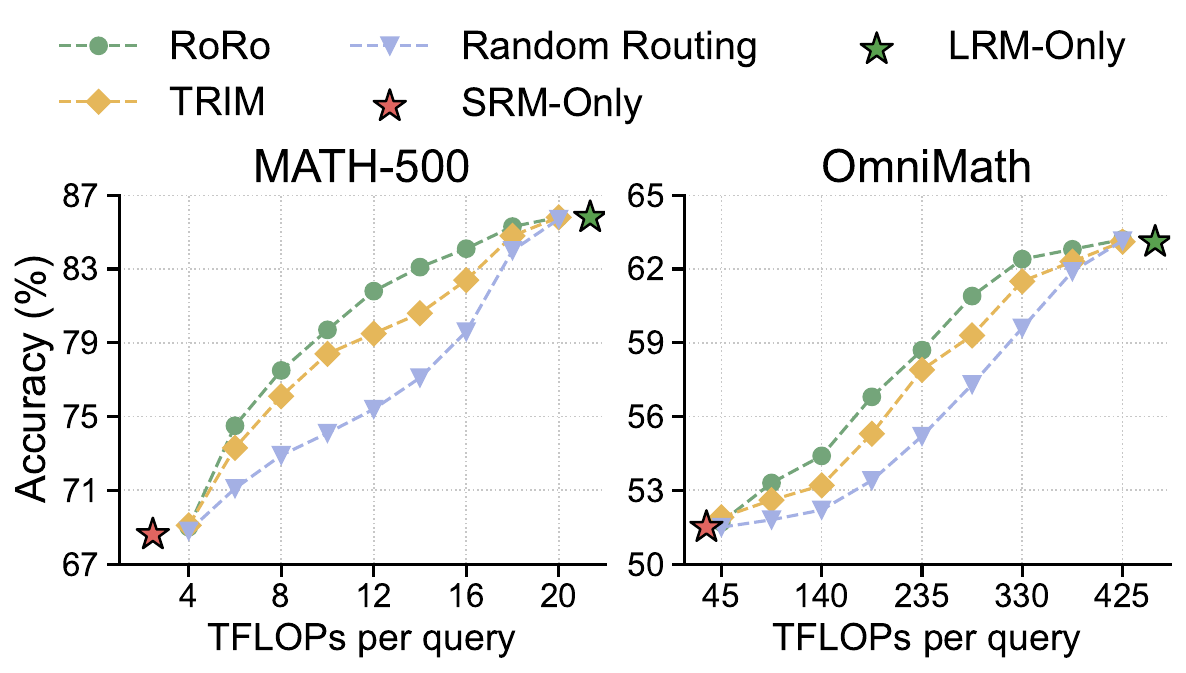}
    \vspace{-6pt}
    \caption{Accuracy-FLOPs trade-off curves under same-family and cross-family settings.}
    \label{fig:trade_off}
    \vspace{-14pt}
\end{figure}

\paragraph{Accuracy--Cost Frontier.}
Figure~\ref{fig:trade_off} shows the accuracy-FLOPs trade-off curves on MATH-500 and OmniMath under both same-family and cross-family settings. Each curve is obtained by sweeping the routing threshold. RoRo achieves a consistently better frontier than all baselines across the full FLOPs range. The advantage is especially clear in the low-budget region, where the router must make precise escalation decisions. In this region, outcome-only methods struggle because sparse reward provides limited guidance for learning effective escalation timing. With rubric-guided process reward, the router receives direct feedback on routing quality, which helps it learn more effective budget allocation under tight constraints.

\paragraph{Latency and Throughput.}
Table~\ref{tab:latency} reports RoRo's end-to-end latency under budget settings. With Qwen3-14B as the LRM, RoRo reduces latency from 104.095s to 41.375s at BA@20, yielding a $2.516\times$ speedup over LRM-only baseline. With Qwen3-32B as the LRM, baseline latency increases to 156.143s due to larger model size, and RoRo achieves larger speedups across budgets. At BA@20, speedup reaches $2.765\times$, and even at BA@60 the speedup ($1.456\times$) remains higher than the 14B counterpart ($1.315\times$). This is expected because the model size gap between SRM and LRM amplifies latency savings from routing to SRM. These results confirm that FLOPs savings translate into wall-clock speedups and that the benefit grows with model size disparity.

\subsection{Further Analysis}
\label{sec:exp_analysis}

\begin{figure}[t]
    \centering
    \includegraphics[width=1\linewidth]{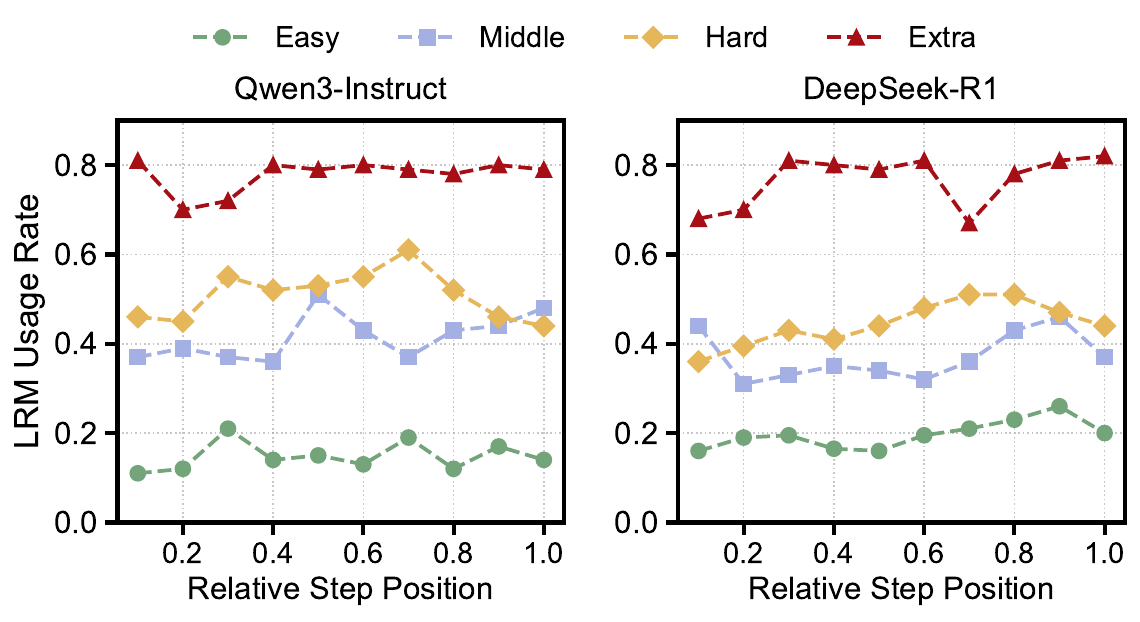}
    \vspace{-19pt}
    \caption{LRM usage rate across OmniMath difficulty levels under same-family and cross-family settings.}
    \label{fig:routing_behavior}
    \vspace{-15pt}
\end{figure}

\paragraph{Difficulty Sensitivity.}
We examine whether RoRo is sensitive to problem difficulty. OmniMath provides human-annotated difficulty levels ranging from 1 to 10 for each problem. We partition these into four groups: \textit{easy} (1--3), \textit{medium} (4--5), \textit{hard} (6--7), and \textit{extra hard} (8--10), and compute the LRM usage rate of each method within each group.
Figure~\ref{fig:routing_behavior} shows the results under both same-family and cross-family settings. Across both settings, RoRo shows a clearer separation across difficulty levels compared with TRIM. It uses the LRM more frequently on hard and extra hard problems and less on easy ones, while TRIM shows a flatter distribution. This indicates that rubric-based process reward helps the router develop a stronger sense of when escalation is truly needed, rather than distributing LRM usage uniformly.

\paragraph{Uncertainty Signal Sensitivity.}
Since the routing policy relies on uncertainty-derived features, we study whether RoRo is sensitive to the choice of uncertainty signal. We replace the default average token entropy with three alternatives and evaluate the resulting routers on MATH-500 and OmniMath. The alternatives include Avg.\ Confidence (average maximum token probability per step), Avg.\ NLL (average negative log probability of generated tokens), and First-3 Token Entropy (average entropy over the first three tokens per step). Table~\ref{tab:uncertainty_sensitivity} reports the results at three budget levels. Among all signals, average entropy achieves the best overall performance, suggesting that full-step entropy provides a more comprehensive measure of step-level uncertainty. Avg.\ NLL yields close results, while Avg.\ Confidence shows a slightly larger gap. First-3 Token Entropy shows the largest drop, likely because the first few tokens carry limited information about overall step difficulty. Nevertheless, all four signals produce competitive results with only minor fluctuations, confirming that RoRo is not tightly coupled to any specific uncertainty signal.

\begin{table}[t]
  \centering
  \small
  \resizebox{\columnwidth}{!}{
  \begin{tabular}{lccc}
  \toprule
  Model Configuration & BA & Latency (s/query) & Speedup \\
  \midrule
  Qwen3-14B & -- 
  & 104.095 & 1.000$\times$ \\
  \rowcolor[HTML]{EDEDED}
  RoRo (1.7B + 14B) & @20 
  & 41.375 & 2.516$\times$ \\
  \rowcolor[HTML]{EDEDED}
  RoRo (1.7B + 14B) & @40 
  & 60.121 & 1.731$\times$ \\
  \rowcolor[HTML]{EDEDED}
  RoRo (1.7B + 14B) & @60 
  & 79.135 & 1.315$\times$ \\
  \midrule
  Qwen3-32B & -- 
  & 156.143 & 1.000$\times$ \\
  \rowcolor[HTML]{EDEDED}
  RoRo (1.7B + 32B) & @20 
  & 56.473 & 2.765$\times$ \\
  \rowcolor[HTML]{EDEDED}
  RoRo (1.7B + 32B) & @40 
  & 78.854 & 1.980$\times$ \\
  \rowcolor[HTML]{EDEDED}
  RoRo (1.7B + 32B) & @60 
  & 107.259 & 1.456$\times$ \\
  \bottomrule
  \end{tabular}
  }
  \vspace{-6pt}
  \caption{End-to-end latency comparison of RoRo. The SRM is Qwen3-1.7B. Speedup is computed relative to the corresponding LRM-only baseline.}
  \label{tab:latency}
  \vspace{-16pt}
  \end{table}

\section{Related Work}
\paragraph{Stepwise Model Routing.}
As large reasoning models increasingly rely on long chain-of-thought trajectories, reducing inference cost without sacrificing reasoning quality has become an important direction. Recent work studies stepwise model routing, where small and large models are dynamically coordinated at the level of intermediate reasoning steps rather than whole queries. Speculative reasoning methods, such as SpecCoT \citep{shi2025speccot} and SpecReason \citep{pan2025specreason}, accelerate reasoning by letting smaller models draft intermediate steps and using stronger models for verification, correction, or fallback. Other methods introduce explicit routing signals, including process reward models in RSD \citep{liao2025rewardguided}, confidence-based routing in STEER \citep{lee2026confidence}, and entropy-based difficulty estimation in GlimpRouter \citep{zeng2026glimprouter}. More recently, TRIM \citep{kapoor2026trim} formulates stepwise routing as a sequential decision problem under accuracy-cost trade-offs, selectively routing critical steps to larger models to prevent cascading failures. However, existing methods mainly rely on local uncertainty, step-level verification, or final outcome-cost trade-offs, leaving the process quality of the routing trajectory itself underexplored.

\paragraph{Rubric-Based RL.}
Rubrics have recently emerged as structured and interpretable supervision for reward modeling and reinforcement learning. Unlike scalar rewards or pairwise preferences, rubrics decompose evaluation into explicit criteria, making them suitable for tasks where correctness is hard to verify. OpenRubrics \citep{liu2025openrubrics} constructs large-scale synthetic rubrics for rubric generation and rubric-based reward modeling, while Rubrics as Rewards \citep{gunjal2026rubrics} uses instance-specific rubric feedback as on-policy reward signals beyond strictly verifiable domains. Recent methods further explore rubrics as dynamic guidance for policy optimization. RuscaRL \citep{zhou2026breakingexplorationbottleneckrubricscaffolded} uses checklist-style rubrics as both rollout scaffolds and reward references, RLCER \citep{sheng2026reinforcing} introduces self-evolving rubrics to provide process supervision for chain-of-thought reasoning, and OpenRS \citep{jia2026open} develops adaptive pairwise rubrics for fine-grained reward supervision. Inspired by these works, we argue that stepwise model routing also requires explicit process-level rubrics to evaluate whether a routing trajectory escalates at appropriate steps, avoids unstable switching, prevents error propagation, and uses computation efficiently.

\begin{table}[t]
\centering
\resizebox{0.48\textwidth}{!}{
\begin{tabular}{l|ccc|ccc}
\toprule
\multirow{2}{*}{\textbf{Uncertainty Signal}} & \multicolumn{3}{c|}{\textbf{MATH-500}} & \multicolumn{3}{c}{\textbf{OmniMath}} \\
\cmidrule(lr){2-4} \cmidrule(lr){5-7}
& @20 & @40 & @60 & @20 & @40 & @60 \\
\midrule
\rowcolor[HTML]{D8ECE4}
Avg.\ Entropy & \textbf{75.5} & \textbf{79.8} & \textbf{81.9} & \textbf{56.5} & \textbf{59.9} & \textbf{62.8} \\
Avg.\ Confidence & 74.9 & 79.0 & 81.2 & 56.5 & 59.7 & 62.9 \\
Avg.\ NLL & 75.2 & 79.5 & 81.6 & 56.2 & 59.6 & 62.5 \\
First-3 Token Entropy & 75.3 & 79.3 & 81.4 & 56.3 & 58.9 & 62.6 \\
\bottomrule
\end{tabular}
}
\vspace{-6pt}
\caption{Uncertainty signal sensitivity analysis on MATH-500 and OmniMath. Avg.\ Entropy (highlighted) is the default signal used in all other experiments.}
\label{tab:uncertainty_sensitivity}
\vspace{-14pt}
\end{table}

\section{Conclusion}
\vspace{-4pt}
In this paper, we study process rewards for stepwise model routing. Existing methods model routing as a sequential process, but still optimize routers mainly with outcome-level rewards. They provide little guidance on whether intermediate routing decisions are well made. To address this gap, we propose RoRo, a rubric-guided process reward framework for router training. RoRo trains a Rubricor and a Judge to generate query-specific rubrics and produce reliable process rewards through alternating optimization and statistical validation.
Experiments on five reasoning benchmarks show that RoRo consistently achieves better accuracy and cost trade-offs than strong baselines under both same-family and cross-family settings. These results show that process-level supervision is important for learning effective routing policies.

\section*{Limitations}
RoRo relies on uncertainty signals to extract state features for the routing policy. Although our experiments show that performance is robust across different uncertainty signal choices, the framework assumes the availability of token-level probability distributions from the SRM, which may not be accessible through all inference APIs. Furthermore, the rubric generation and validation stages introduce training-time cost. While this cost is amortized over many queries and does not affect inference, it may limit applicability in settings where training resources are constrained. Finally, our experiments focus on mathematical reasoning benchmarks. Extending RoRo to broader reasoning domains such as code generation and scientific question answering remains an open direction for future work.

\vspace{-4pt}

\section*{Ethics Statement}

Our work aims to improve the efficiency and reliability of multi-step reasoning through stepwise model routing. However, like any system built on LLMs, the routing framework may still produce incorrect intermediate reasoning steps or factually incorrect final answers. We therefore encourage users to exercise caution and verify critical outputs when deploying such systems in real-world scenarios. Furthermore, our research builds upon open-source models and frameworks, including Qwen3, DeepSeek-R1-Distill, PyTorch, and Hugging Face. We strictly follow their respective licenses and usage policies, and acknowledge their important contributions to the research community.

\section*{Acknowledgement}

This work is supported by the Fundamental Research Funds for the Central Universities under Grant No. WK2102026004 and Anhui Provincial Science and Technology Innovation Key Research Program (202523j08050016).

\vspace{-8pt}

\bibliography{custom}

\appendix

\section{Additional Experimental Setups}
\label{sec:appendix1}

\subsection{Model Configurations}
\label{sec:a1}

Table~\ref{tab:model_config} summarizes the key model and training configurations used in our experiments.

\begin{table}[h]
\centering
\small
\begin{tabular}{ll}
\toprule
\textbf{Component} & \textbf{Configuration} \\
\midrule
SRM & Qwen3-1.7B \\
LRM (same-family) & Qwen3-14B \\
LRM (cross-family) & DeepSeek-R1-Distill-Qwen-14B \\
Rubricor backbone & Qwen3-8B \\
Judge backbone & Qwen3-8B \\
Router architecture & 2-layer MLP, 128 hidden units \\
Router input dim & 5 \\
\midrule
Training data & MATH training set \\
GRPO rollout group $K$ & 8 \\
Cost weight $\lambda$ & 0.001 \\
Process reward weight $\beta$ & 0.5 \\
Temperature & 0.7 \\
Max generation length & 16{,}384 tokens \\
Top-$p$ & 0.95 \\
\bottomrule
\end{tabular}
\caption{Model and training configurations for RoRo.}
\label{tab:model_config}
\end{table}

We use Qwen3-1.7B \citep{yang2025qwen3} as the SRM. For the LRM, we consider Qwen3-14B \citep{yang2025qwen3} for same-family collaboration and DeepSeek-R1-Distill-Qwen-14B \citep{guo2025deepseek} for cross-family collaboration. The routing policy is a 2-layer MLP with 128 hidden units, following TRIM \citep{kapoor2026trim}. It takes a 5-dimensional state vector as input, including three uncertainty-derived features (the current-step uncertainty, the minimum and average uncertainty over the prefix), the step token count normalized by a fixed constant, and the step index normalized by the maximum step limit. By default, we use average token entropy as the uncertainty signal. For the Rubricor and the Judge, we use Qwen3-8B \citep{yang2025qwen3} as the backbone. Both are trained through alternating optimization on the MATH training set and frozen during router training and inference. The cost weight $\lambda = 0.001$ controls the penalty for LRM token usage, and the process reward weight $\beta = 0.5$ balances process supervision against outcome reward. We set $\lambda$ based on the typical LRM token count range so that the cost term is on a comparable scale with the binary outcome reward. We set $\beta$ by grid search over $\{0.1, 0.3, 0.5, 0.7, 1.0\}$ on a held-out validation split of MATH and select 0.5 as it yields the best accuracy-cost trade-off. For all methods, we use the same decoding configuration for fair comparison. All reported results are averaged over three independent runs.

\begin{table*}[t]
    \centering
    \small
    \resizebox{\linewidth}{!}{%
    \begin{tabular}{
        >{\centering\arraybackslash}m{0.16\linewidth}|
        >{\small\arraybackslash}m{0.40\linewidth}|
        >{\small\arraybackslash}m{0.40\linewidth}
    }
    \specialrule{1.2pt}{0pt}{0pt} 
    \textbf{Seed Criterion} & \textbf{Criterion} & \textbf{Purpose} \\
    \midrule
    
    \textbf{Seed 1} 
    &
    The route should switch to the LRM near the first SRM step that is likely to cause, or has just caused, a critical reasoning error, based on both the step content and its difficulty signals.
    &
    This criterion evaluates whether the router can perform timely LRM intervention near the first SRM failure point by jointly considering the reasoning content, step-level difficulty signals, and routing actions.
    \\
    
    \midrule
    
    \textbf{Seed 2} 
    &
    The route should use the LRM to repair wrong, unstable, or misleading intermediate states produced by the SRM, and the later trajectory should reflect this recovery.
    &
    This criterion evaluates whether LRM intervention actually recovers the reasoning trajectory after SRM errors, rather than merely increasing model cost without improving the reasoning state.
    \\
    
    \midrule
    
    \textbf{Seed 3}
    &
    The route should allocate LRM calls only to steps where the expected reasoning benefit justifies the additional cost, while keeping easy or low-risk steps on the SRM.
    &
    This criterion evaluates whether the router uses LRM computation selectively, based on step difficulty, step importance, routing actions, and the actual cost of each model call.
    \\
    
    \specialrule{1.2pt}{0pt}{0pt} 
    \end{tabular}
    }
    \caption{The seed rubric used by RoRo. The rubric consists of three routing-specific criteria that evaluate the quality of a complete routing trajectory from step content, step-level difficulty signals, routing actions, and model-call costs.}
    \label{tab:seed_rubric}
    \vspace{-10pt}
\end{table*}

\subsection{Dataset Details}
\label{sec:a2}

\paragraph{MATH-500 \citep{lightman2023let}.}
MATH-500 is a curated subset of 500 problems sampled from the MATH benchmark \citep{hendrycks2021measuring}, covering seven mathematical subjects including algebra, number theory, geometry, and combinatorics. The problems span five difficulty levels, with Level 5 requiring multi-step reasoning and creative problem solving. MATH-500 is widely adopted for evaluating reasoning capabilities of LLMs.

\vspace{-4pt}

\paragraph{AIME 2025 \citep{dekoninck2026beyond}.}
AIME 2025 contains 30 competition-level problems from the 2025 American Invitational Mathematics Examination. Each problem requires integer answers between 0 and 999. AIME problems are significantly more challenging than typical benchmarks and demand advanced mathematical reasoning, including complex algebraic manipulation, number theory, and geometric reasoning.

\vspace{-4pt}

\paragraph{OmniMath \citep{gao2025omnimath}.}
OmniMath is a large-scale benchmark of 4,428 competition-level mathematics problems sourced from international olympiads and national contests. Each problem is annotated with a human difficulty rating from 1 to 10. The benchmark spans a wide range of mathematical topics and requires advanced problem-solving strategies. We use the subset with difficulty levels 1--10 for our evaluation.

\vspace{-4pt}

\paragraph{GSM8K \citep{cobbe2021training}.}
GSM8K is a benchmark of 8,792 grade-school math word problems that require multi-step arithmetic reasoning. Although the individual reasoning steps are simple, the problems require 2--8 steps of basic arithmetic operations. We use GSM8K as an out-of-domain benchmark to test whether routers trained on MATH generalize to simpler problem distributions.

\vspace{-4pt}

\paragraph{GPQA \citep{rein2024gpqa}.}
GPQA (Graduate-level Google-Proof QA) is a benchmark of 448 challenging multiple-choice questions in biology, physics, and chemistry. The questions are designed by domain experts to be answerable by experts but difficult for non-experts, even with access to search engines. We use GPQA as an out-of-domain benchmark to evaluate whether routing strategies generalize from mathematical reasoning to scientific reasoning.

\subsection{Seed Rubric}
\label{sec:appendix_rubrics}

The seed rubric provides the initial process-level supervision for training the Judge. It consists of three manually designed criteria, each targeting a distinct aspect of routing quality. Table~\ref{tab:seed_rubric} summarizes all three criteria along with their purposes.

\textbf{Seed 1} focuses on timely escalation. A good routing trajectory should switch to the LRM near the first SRM step that is likely to produce a critical reasoning error. This criterion jointly considers the step content, step-level difficulty signals, and the routing action to evaluate whether the escalation happens at an appropriate point.

\textbf{Seed 2} focuses on effective recovery. Simply invoking the LRM is not sufficient if the routing trajectory does not improve afterward. This criterion evaluates whether the LRM intervention actually repairs erroneous or unstable intermediate states produced by the SRM, and whether the later trajectory reflects this recovery.

\textbf{Seed 3} focuses on cost efficiency. The router should allocate LRM calls only to steps where the expected reasoning benefit justifies the additional cost. Easy or low-risk steps should remain on the SRM. This criterion evaluates whether the routing trajectory uses computation selectively based on step difficulty and importance.

Together, these three criteria capture the key trade-offs in stepwise model routing. They serve two roles in RoRo. First, they provide initial process preferences for constructing the preference dataset $\mathcal{D}_q$. Second, they condition the Judge during warm-start training. After alternating optimization, the learned Rubricor generates query-specific criteria that go beyond the seed rubric and adapt to different problem types.

\begin{table*}[t]
\centering
\small
\setlength{\tabcolsep}{6pt}
\renewcommand{\arraystretch}{1.12}
\begin{tabular}{l l l c c c c}
\specialrule{1.2pt}{0pt}{0pt}
\textbf{Case} & \textbf{Source} & \textbf{Method} 
& \textbf{LRM Calls} & \textbf{LRM Tokens} & \textbf{TFLOPs} & \textbf{Correct} \\
\midrule

\multirow{3}{*}{Case 1} 
& \multirow{3}{*}{\texttt{omnimath\_01261}} 
& SRM-only & 0 / 30 & 0 & -- & \xmark \\
& & TRIM & 24 / 30 & 10{,}223 & 346.57 & \xmark \\
\rowcolor[HTML]{D8ECE4}
& & RoRo & 13 / 30 & 5{,}777 & 222.09 & \cmark \\

\midrule

\multirow{3}{*}{Case 2} 
& \multirow{3}{*}{\texttt{omnimath\_01924}} 
& SRM-only & 0 / 30 & 0 & -- & \xmark \\
& & TRIM & 21 / 30 & 4{,}824 & 141.81 & \xmark \\
\rowcolor[HTML]{D8ECE4}
& & RoRo & 10 / 30 & 2{,}244 & 71.37 & \cmark \\

\specialrule{1.2pt}{0pt}{0pt}
\end{tabular}
\vspace{4pt}
\caption{Routing statistics and outcomes for two OmniMath problems. TRIM invokes the LRM on most steps but still produces wrong answers on both problems. RoRo uses significantly fewer LRM calls and lower FLOPs while reaching the correct answer. \cmark\ and \xmark\ denote correct and incorrect answers.}
\label{tab:case-study-combined}
\end{table*}

\section{Additional Implementation Details}
\label{sec:appendix_validation}

\subsection{Validation Gate Details}
\label{sec:validation_gate}

This section provides full details of the validation gate used in rubric filtering (Sec.~\ref{sec:rubric_reward_modeling}). The validation gate filters generated rubric criteria to retain only those that are genuinely predictive of routing quality, while avoiding outcome leakage and redundancy. For each criterion $c_i$ in a candidate rubric, we compute its score on a held-out set of routing rollouts and apply three statistical tests. A criterion must pass all three tests to be retained.

\paragraph{Held-Out Rollout Set.}
We reserve 200 queries from the MATH training set as a held-out validation split. For each query, we collect routing rollouts from the same six policies described in Sec.~\ref{sec:trajectory_collection}, yielding approximately 1{,}200 rollouts in total. Each rollout is annotated with its final-answer correctness, LRM token usage, and per-step routing actions. This set is fixed throughout training and is used only for criterion validation.

\paragraph{Criterion Scoring.}
Given a criterion $c_i$ and a rollout $\tau$, we prompt the Judge backbone (Qwen3-8B) to assign a score $v_i(\tau) \in [0, 1]$ indicating how well the routing trajectory satisfies the criterion. To reduce variance, we average over two independent Judge calls per rollout. The Judge prompt contains the query, the full routing trajectory, and the criterion text.

\paragraph{Test 1: Partial Correlation Significance.}
We compute the partial correlation between the criterion score $v_i$ and the route preference label (preferred vs.\ dispreferred), controlling for outcome correctness and normalized cost. We use the standard partial Pearson correlation with a two-sided $t$-test. A criterion is retained only if $p < 0.05$ after Holm-Bonferroni correction for multiple testing across all criteria in the candidate rubric.

\paragraph{Test 2: Score Variance.}
We require that the standard deviation of criterion scores across rollouts exceeds $\sigma_{\min} = 0.05$. Criteria with near-constant scores across trajectories carry no discriminative information and are removed.

\paragraph{Test 3: Outcome Leakage.}
We compute the mutual information $\text{MI}(v_i; y)$ between the criterion score and the binary outcome label $y \in \{0, 1\}$. A criterion is removed if $\text{MI}(v_i; y) > 0.1$ nats. This threshold ensures that retained criteria capture routing process quality rather than simply reflecting whether the final answer is correct. We estimate mutual information using the KSG estimator \citep{kraskov2004estimating} with $k = 5$ neighbors.

\paragraph{Multiple Testing Correction.}
When a candidate rubric contains $|r|$ criteria, Test 1 involves $|r|$ simultaneous hypothesis tests. We apply Holm-Bonferroni correction to control the family-wise error rate at $\alpha = 0.05$. Criteria that fail the corrected significance threshold are removed regardless of their variance or leakage properties.

\paragraph{Rubric Discard Policy.}
If fewer than two criteria pass all three tests, the entire candidate rubric is discarded. During Rubricor training, a discarded rubric receives zero reward. During router training, if the generated rubric for a GRPO rollout group fails validation, the entire rollout group is excluded from that training step. In our experiments, the discard rate is approximately 12\% during Rubricor training and drops below 5\% by the final alternating round, indicating that the Rubricor learns to generate valid criteria over time.

\subsection{Prompt}
\label{sec:prompt_templates}

Table~\ref{tab:prompt_rubricor} shows the prompt template used by the Rubricor to generate routing-quality criteria. Table~\ref{tab:prompt_judge} shows the prompt template used by the Judge to score a routing trajectory under a given rubric.

\begin{table*}[ht]
  \small
  \centering
  \begin{tabularx}{\textwidth}{>{\ttfamily\raggedright\arraybackslash}X}
    \hline
    \textbf{Prompt of Rubricor}\tabularnewline
    \hline
    \#\#\# System:\newline
    You are the Rubricor in a stepwise model routing system for reasoning. A weak reasoning model (SRM) proposes each reasoning step. A router either accepts the SRM step or replaces it with a strong reasoning model step. The goal is to preserve reasoning quality while avoiding unnecessary LRM calls.\newline
    You will be given one problem and a pool of routing trajectories for that problem. Each trajectory contains the selected reasoning steps and the model used at each step.\newline
    \#\#\# Input:\newline
    [Question]: \{question\}\newline
    [Trajectory 1]: \{trajectory\_1\}\newline
    [Trajectory 2]: \{trajectory\_2\}\newline
    ...\newline
    [Trajectory N]: \{trajectory\_N\}\newline
    The input does not provide final correctness labels or preference labels. Do not assume that any trajectory is preferred.\newline
    \#\#\# Task:\newline
    Generate 3--5 general routing-quality criteria for evaluating routing trajectories under this question. Each criterion must satisfy all requirements:\newline
    1. It evaluates the routing process rather than final answer correctness.\newline
    2. It is label-agnostic: it must not refer to trajectory IDs, final correctness, reference answers, or which trajectory is preferred.\newline
    3. It should be applicable beyond this specific problem, while still being relevant to the routing challenges shown in the trajectory pool.\newline
    4. It should capture whether the route prevents, repairs, or verifies high-impact reasoning errors.\newline
    5. It must not collapse into trivial heuristics such as "always use LRM in later steps", "always minimize LRM calls", "always avoid switching", or "use LRM whenever the solution is long".\newline
    \#\#\# Possible Aspects:\newline
    Possible aspects include, but are not limited to: intervention before error propagation; timeliness of escalation under uncertainty; avoiding LRM calls on routine or already reliable steps; recovery after contradiction or uncertainty.\newline
    \#\#\# Output Format:\newline
    Return only a JSON object:\newline
    \{"rubrics": [\{"criterion": "one sentence describing what the route should do or should not do", "score": 0.8, "weight": 0.25\}]\}\newline
    Each weight must be a number in [0,1].\tabularnewline
    \hline
  \end{tabularx}
  \caption{Prompt template for the Rubricor.}
  \label{tab:prompt_rubricor}
\end{table*}

\begin{table*}[ht]
  \small
  \centering
  \begin{tabularx}{\textwidth}{>{\ttfamily\raggedright\arraybackslash}X}
    \hline
    \textbf{Prompt of Judge}\tabularnewline
    \hline
    \#\#\# System:\newline
    You are the Judge in a stepwise model routing system for reasoning. A weak reasoning model (SRM) proposes each reasoning step. A router either accepts the SRM step or replaces it with a strong reasoning model step (LRM). The goal is to preserve reasoning quality while avoiding unnecessary LRM calls.\newline
    You will be given one problem, one routing trajectory, and a rubric generated by the Rubricor. Your task is to decide whether the trajectory satisfies each rubric criterion, and then compute a weighted process score.\newline
    \#\#\# Input:\newline
    [Question]: \{question\}\newline
    [Routing Trajectory]: \{trajectory\}\newline
    [Rubric]: \{rubric\_json\}\newline
    The trajectory contains the selected reasoning steps and the model used at each step. It may include a final answer in the text, but you must not judge whether the final answer is correct. You should evaluate only the routing process.\newline
    \#\#\# Task:\newline
    For each rubric criterion:\newline
    1. Decide whether the trajectory satisfies the criterion.\newline
    2. Set "satisfied" to true if the routing behavior clearly satisfies the criterion.\newline
    3. Set "satisfied" to false if the routing behavior clearly violates the criterion or lacks evidence for satisfying it.\newline
    4. Use the criterion and score\_guidance to make the decision.\newline
    Compute the final process score as:\newline
    final\_score = sum of weight * score * indicator\newline
    where indicator = 1 if satisfied is true and 0 otherwise.\newline
    \#\#\# Output Format:\newline
    Return only a valid JSON object:\newline
    \{"criterion\_judgments": [\{"criterion": "the original criterion text", "score": 0.5, "satisfied": true\} ...], "final\_score": 0.0\}\tabularnewline
    \hline
  \end{tabularx}
  \caption{Prompt template for the Judge.}
  \label{tab:prompt_judge}
\end{table*}

\section{Additional Related Work}
\label{sec:appendix_related}

\paragraph{Adaptive and Efficient LLM Reasoning.}
Large language models have demonstrated strong reasoning capabilities across diverse tasks and application domains, motivating extensive studies on their evaluation, adaptation, and robustness~\cite{chang2024survey}. Recent studies have explored optimization modeling~\cite{liu2026automated}, structured knowledge and graph reasoning~\cite{li-etal-2025-enrich,li2026layerlogitslogicempowering}, formal reasoning~\cite{chen2024llmsprobabilisticminimallyadequate}, software analysis~\cite{zhang2024glitchprober}, and multimodal understanding and generation~\cite{zhang2026temporal,tang2026efficientpostergen,xiao2026staying}. LLMs are also increasingly used in agentic systems involving tool use, GUI interaction, long-horizon memory, and multimodal search~\cite{liu2026toolanchor,shi2026androtmem,yu2026lightweight,yu2026cirrusbench,chen2026learningsearchdecisionbasedagent,wu2026promsa,hao2026evolve,liu2026meta}. Beyond improving reasoning through fine-tuning, parameter-efficient adaptation, reinforcement learning, and knowledge augmentation~\cite{chang2026balora,xiao2026not,liu2026auditingdatamembershipreinforcement,lv2026pcsdpersistentconsistencyselfdistillation,wu2026contrastive,li2026scaling,hao2024towards,hao2026rethinking,wang2026scheduling}, recent work has investigated adaptive reasoning behaviors through structural robustness analysis~\cite{he2026order}, step-level reasoning optimization~\cite{shi2026spader}, and efficient reasoning process control~\cite{liu2025thoughtmanipulationexternalthought,zhou2026look,zhou2026one}.
In parallel, substantial efforts have been devoted to improving the efficiency of LLM inference. Existing approaches reduce computational and memory costs through model quantization~\cite{zhao2025quark,zhao2026specquant,xu2026kbvq,zhao2026twla}, parameter-efficient adaptation and model optimization~\cite{NEURIPS2025_6724eae9}, sparse expert pruning and reorganization~\cite{zhou2025dropping}, or adaptive compression of redundant multimodal tokens~\cite{tang2026codeshrink,tang2026efficientpostergen}. These methods mainly improve efficiency by reducing the cost of executing a fixed model or processing a fixed input. Model routing instead exploits heterogeneous capability--cost trade-offs among different models and dynamically selects suitable models for inference. Our work further studies \emph{stepwise model routing}, which operates at a finer granularity by allowing model allocation to change within a single reasoning trajectory, thereby adapting computational resources to the requirements of different reasoning steps.

\paragraph{Process Reward Modeling.}
Process reward modeling provides dense supervision for multi-step reasoning beyond final-answer correctness. Existing studies propose step-level reward models that provide intermediate navigation signals for reasoning chains \citep{ma2023letsrewardstepstep}, scale automated process verifiers to provide dense verification for each reasoning step \citep{setlur2024rewardingprogressscalingautomated}, and introduce stepwise process-aware rewards that do not rely on reference answers, enabling broader applicability to tasks without gold intermediate labels \citep{rahman2025spark}. More recently, rubric-based rewards with stepwise attribution have been studied for credit assignment in LLM reasoning \citep{xie2026step}. These works demonstrate the value of dense process-level supervision for reasoning tasks. Our work extends this line of research to routing trajectories, where process quality is even harder to define and verify due to the absence of gold routing labels.

\section{Case Study}
\label{sec:appendix_case}

To provide a more concrete understanding of how RoRo improves routing quality, we present two representative case studies from OmniMath. In both cases, the SRM-only baseline produces incorrect answers, TRIM assigns most steps to the LRM but still fails, while RoRo selects fewer LRM steps and reaches the correct answer with lower cost. Table~\ref{tab:case-study-combined} summarizes the routing statistics and outcomes for both cases.

\begin{figure}[t]
    \centering
    \includegraphics[width=1\linewidth]{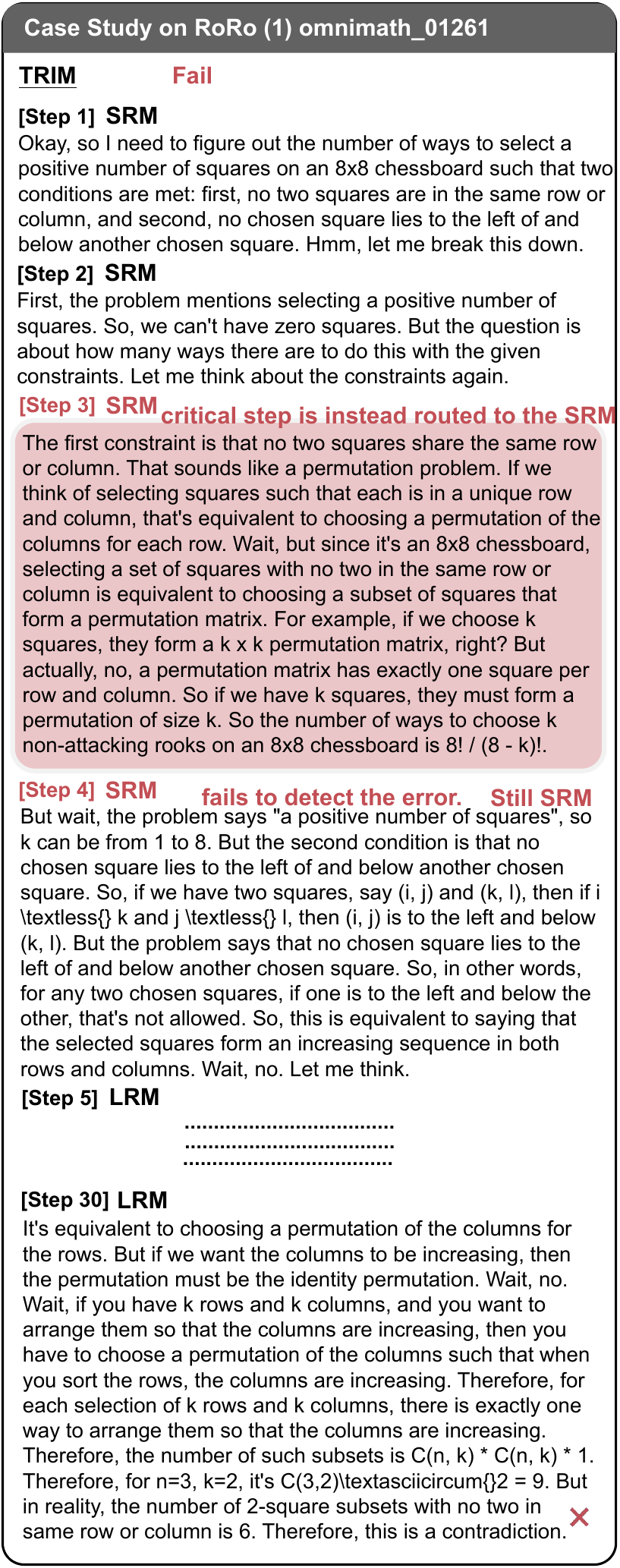}
    \caption{Routing trajectory of TRIM on Case 1 (\texttt{omnimath\_01261}). TRIM routes 24 out of 30 steps to the LRM but misses the early critical steps, leading to an incorrect answer.}
    \label{fig:case1}
\end{figure}

\begin{figure}[t]
    \centering
    \includegraphics[width=1\linewidth]{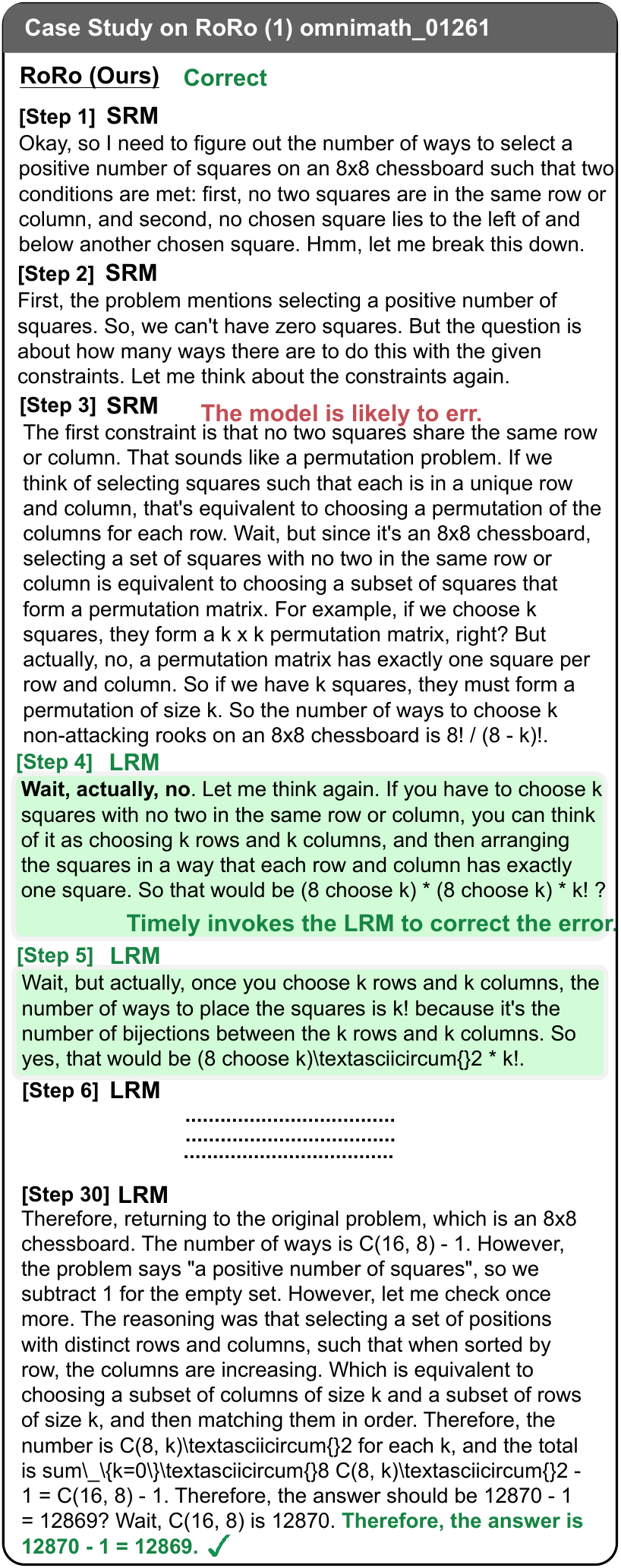}
    \caption{Routing trajectory of RoRo on Case 1 (\texttt{omnimath\_01261}). RoRo concentrates LRM calls on the first 9 steps for problem formulation and reaches the correct answer with 13 LRM calls.}
    \label{fig:case2}
\end{figure}

\paragraph{Case 1 (omnimath\_01261).}
This problem asks for the number of ways to select squares on an $8\times 8$ chessboard under row, column, and monotonicity constraints (gold answer: 12{,}869). As shown in Figures~\ref{fig:case1} and \ref{fig:case2}, TRIM routes 24 out of 30 steps to the LRM. Despite this heavy usage, TRIM fails to allocate the LRM to the early critical steps where the SRM makes a wrong combinatorial formulation. The error propagates through the rest of the trajectory and leads to an incorrect answer (1{,}441{,}728). In contrast, RoRo concentrates LRM calls on the first 9 steps, where the problem formulation and mathematical setup are most critical. After the reasoning direction is established correctly, RoRo switches back to the SRM for the remaining computation. This results in a correct answer with only 13 LRM calls and 36\% fewer FLOPs than TRIM.

\begin{figure}[t]
    \centering
    \includegraphics[width=1\linewidth]{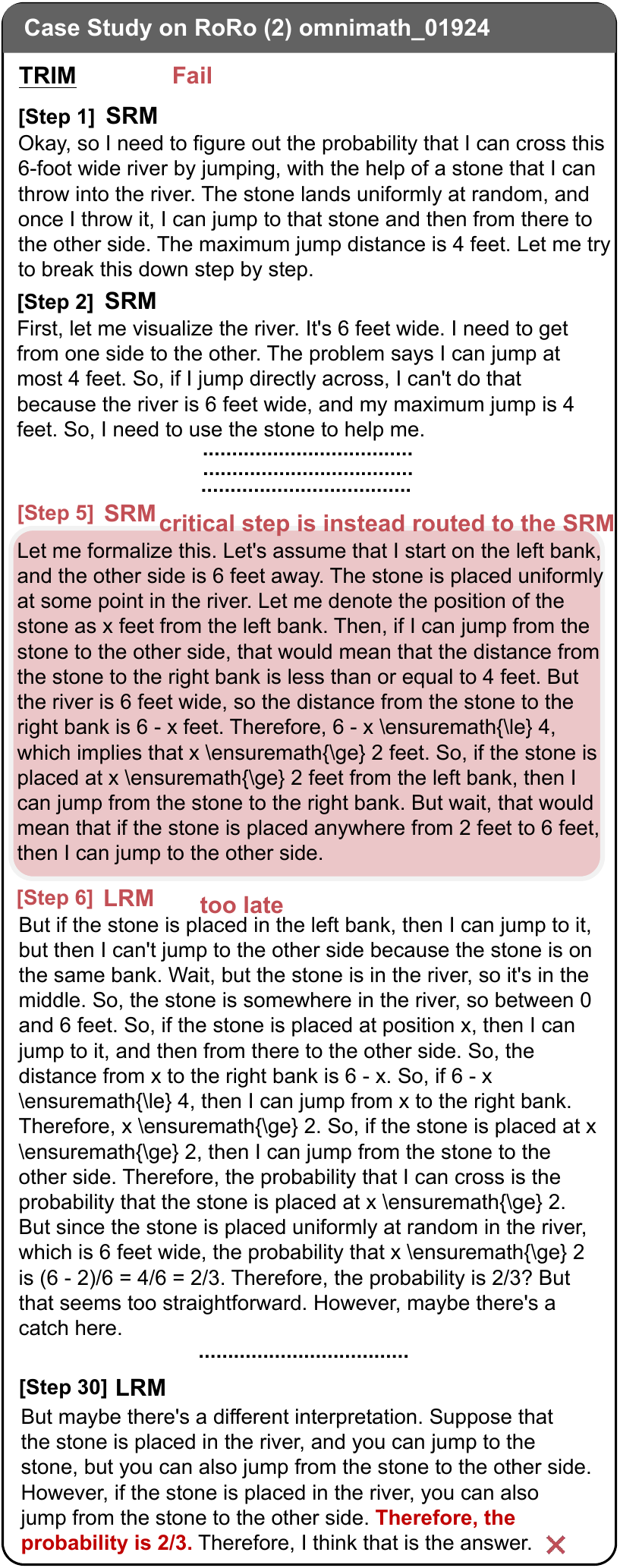}
    \caption{Routing trajectory of TRIM on Case 2 (\texttt{omnimath\_01924}). TRIM invokes the LRM mainly in later steps and fails to correct the early SRM error in probability setup.}
    \label{fig:case3}
\end{figure}

\begin{figure}[t]
    \centering
    \includegraphics[width=1\linewidth]{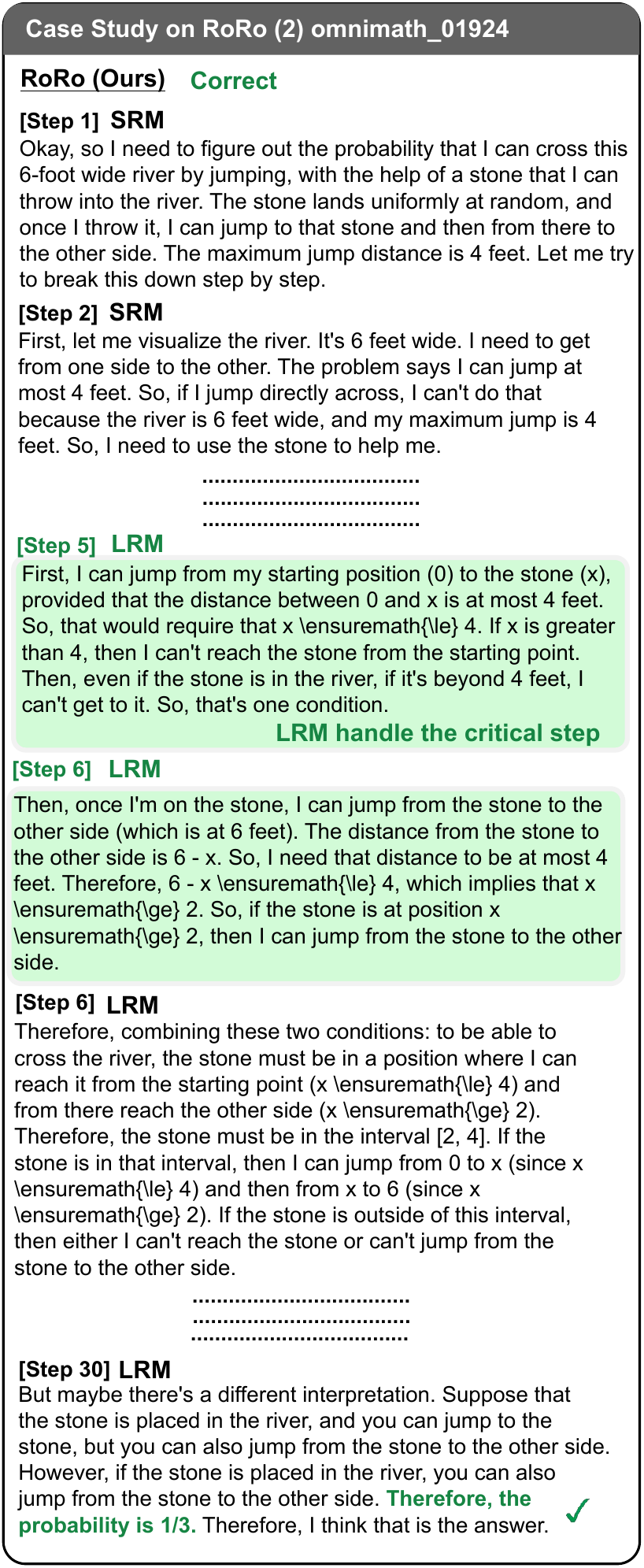}
    \caption{Routing trajectory of RoRo on Case 2 (\texttt{omnimath\_01924}). RoRo invokes the LRM at the early critical stage and reaches the correct answer with only 10 LRM calls.}
    \label{fig:case4}
\end{figure}

\paragraph{Case 2 (omnimath\_01924).}
This problem involves computing the probability of crossing a river by throwing a stone uniformly at random (gold answer: $\frac{1}{3}$). Figures~\ref{fig:case3} and \ref{fig:case4} illustrate the routing trajectories. TRIM again assigns most steps (21 out of 30) to the LRM, but the LRM is invoked too late. The SRM has already committed to an incorrect probability setup in the early steps, and later LRM calls cannot recover from this error. RoRo, on the other hand, invokes the LRM at an earlier stage to correctly set up the integral formulation, and then returns to the SRM once the reasoning is on the right track. With only 10 LRM calls, RoRo reaches the correct answer while using roughly half the FLOPs.

These two cases illustrate a key advantage of RoRo over TRIM. TRIM tends to distribute LRM calls uniformly or toward the later steps, missing early critical steps where SRM errors are most consequential. RoRo, guided by rubric-based process rewards, learns to place LRM calls at the steps where escalation brings the most benefit, leading to both higher accuracy and lower cost.

\end{document}